%% file: ieee.tex
\documentclass[conference]{IEEEtran}
\IEEEoverridecommandlockouts
\usepackage{cite}
\usepackage{amsmath,amssymb,amsfonts}
\usepackage{algorithmic}
\usepackage{graphicx}
\usepackage{textcomp}
\usepackage{xcolor}

\usepackage[export]{adjustbox}

\usepackage{url}
\usepackage[caption=false,font=footnotesize,labelfont=sf,textfont=sf]{subfig}

\begin{document}

\title{YouTube AV 50K: An Annotated Corpus for Comments in Autonomous Vehicles
}

% \and
% \and
% \and
% \and
% \and

\author{Tao Li, Lei Lin, Minsoo Choi, Kaiming Fu, Siyuan Gong, Jian Wang \\
Purdue University \\
\texttt{\{taoli,lin954,choi502,fu241,gong131,wang2084\}@purdue.edu}
}

\maketitle

\input{content}

\bibliography{db}
\bibliographystyle{IEEEtran}

\end{document}

%% file: content.tex
\begin{abstract}
With one billion monthly viewers, and millions of users discussing and sharing opinions, comments below YouTube videos are rich sources of data for opinion mining and sentiment analysis. We introduce the YouTube AV 50K dataset, a freely-available collections of more than 50,000 YouTube comments and metadata below autonomous vehicle (AV)-related videos. We describe its creation process, its content and data format, and discuss its possible usages. Especially, we do a case study of the first self-driving car fatality to evaluate the dataset, and show how we can use this dataset to better understand public attitudes toward self-driving cars and public reactions to the accident. Future developments of the dataset are also discussed.
\end{abstract}

\section{Introduction}
Social media has become prevalent and important for social networking and opinion sharing in recent years \cite{asur2010predicting}. By changing the way we perceive and interact with the world, social media has changed our lives profoundly \cite{safko2010social, he2013social}. With millions of posts and replies uploaded every day on social media such as Facebook, Twitters and YouTube, it is an abundant and informative data source of public opinions;
thus, it has attracted lots of attention from both academia and industry to understand people and society \cite{perrin2015social, boulianne2015social, ceron2014every}. Most previous text mining-based social media analysis focused on Twitter and Facebook \cite{kulkarni2018extensive}. YouTube, generally considered as a video platform, the values of its text comments below videos have long been underestimated. Being the second most popular website in the world \cite{alexa2018top} and having 1.9 billion active users \cite{statista2018youtube}, YouTube is an attractive source of research in social media analysis with immense potentials.

Recent developments in autonomous vehicle technology have helped bring self-driving vehicles to the forefront of public interest \cite{schoettle2014survey}. Autonomous vehicles, particularily after the first fatal crash of self-driving cars recently happened in Arizona \cite{wakabayashi2018self}, have become a popular topic in social media. In order to investigate users' acceptance, safety concerns, and willingness to purchase of autonomous vehicles, extensive studies have been conducted \cite{howard2014public, kyriakidis2015public, schoettle2014public}. However, these research rely heavily on surveys, which have disadvantages that (i) securing a high response rate is hard; (ii) uncertainty over the validity of the data and sampling issues; and (iii) concerns surrounding the design, implementation, and evaluation of the survey \cite{wright2005researching, kelley2003good, fowler2013survey, fricker2002advantages}. Latest techniques in opinion mining and sentiment analysis from social media data  \cite{pang2008opinion, pak2010twitter, esuli2007sentiwordnet, baccianella2010sentiwordnet, liu2012sentiment} offer new possibilities to overcome disadvantages of traditional surveys, and the YouTube AV 50K dataset is introduced under this background.

This paper is organized as follows. In Section \ref{sec:dataset} we introduce our motivation of building this dataset as well as the data creation process, annotation methods, data formats, and basic statistics of the dataset. We propose possible usages of the dataset with examples in Section \ref{sec:usage}, including metadata analysis, text visualization, sentimental analysis, recommendation systems, and text regression. Related works are presented in Section \ref{sec:related}. Finally, Section \ref{sec:future} describes our visions and future goals of this project.

\section{The Dataset}\label{sec:dataset}

\subsection{Motivation}
The YouTube Autonomous Vehicle (AV) 50K dataset is our attempt to help researchers in both natural language processing (NLP) community and autonomous vehicle community by providing a large-scale annotated corpus in the autonomous vehicle domain specifically. The dataset contains comments that are legally available to the public below a comprehensive list of autonomous vehicle-related vedieos on YouTube. Main purposes of the dataset includes: (i) encourage researchers on theories and algorithms to connect their works with real-world data; (ii) provide a reference dataset and benchmarks for evaluating research; and (iii) help new researchers quickly get started in natural langauge processing, data analysis, and autonomous vehicles.

\subsection{Corpus}
The core of the corpus is from the YouTube Data APIs\footnote{\url{https://developers.google.com/youtube/v3/}}. The APIs can be used to search for videos matching specific search terms, topics, locations, publication dates, and much more. Other helpful features for social media analysis include metadata of video, comment, playlist, author, and channel.
For example, original text, like count, publish date, author name, and viewer rating can be found in each comment representation. Basically, any public available resources on YouTube can be downloaded using the APIs. Tutorials and more information can be found in the API document\footnote{\url{https://developers.google.com/youtube/v3/docs/}}.

\subsection{Annotation}
Sentiment score is a numerical representation of the sentiment polarity, the degree of negative, neutral, or positive, for a piece of text. We used the Natural Language API\footnote{\url{https://cloud.google.com/natural-language/}} for sentimental analysis. We consider the sentiment score given by the API as a benchmark and will discuss different models for sentimental analysis in our upcoming paper. Evaluation of the annotation will be shown in the following Section \ref{sec:usage}.

\subsection{Content}
The YouTube AV 50K contains comments and metadata for all public self-driving-related videos on YouTube. Basic statistics is shown in Table \ref{tab:stat} (the numbers might change as the dataset is constantly updating).
\input{table_stat}

The data is stored using JSON format \cite{crockford2006application} described by YouTube Data API to efficiently handle the heterogeneous types of information such as text, user id, publish date, like count, etc. Each comment is described by a JSON object\footnote{\url{https://developers.google.com/youtube/v3/docs/comments}}, whose structure is shown in Figure \ref{fig:comment_json}.

The YouTube AV 50K website\footnote{\url{https://youtubeav50k.goodata.org}} is a core component of the corpus. It contains tutorials, code samples, API documents, an issue section, and the pointers to the actual data, hosted by Goodata Foundation\footnote{\url{https://goodata.org}}.

\input{figure_comment_json}

\section{Proposed Usages}\label{sec:usage}
A wide range of social media analysis tasks can be performed and measured on the YouTube AV 50K dataset. In this section, we propose some possible usages of the dataset, and illustrate and evaluate the power of the dataset.

\input{figure_meta_histgram}
\subsection{Metadata Analysis}\label{sec:metadata}
A basic usage of the dataset is to analyze the metadata. This task is already very interesting, because trending topics, geographic information, and even patterns of human behaviors can be discovered from the metadata of a large volume of comments and videos. Statistical descriptions and visualisations can also be done for related research. An example of analyzing human comment behaviors is given in Figure \ref{fig:meta_histgram}, which illustrates a clear pattern that YouTube users are most active at around 8pm and least active in the early morning. Suprisingly, the figure shows that YouTube users leave more comments in average on every weekday rather than weekends. Another example usage is to provide channel rankings using the video metadata from the dataset. We provide a ranking of top 20 English news channels in YouTube by their monthly views from May 9, 2018 to June 9, 2018 in Table \ref{table_top20}. The results are exactly the same compared to data from KEDOO\footnote{\url{https://www.kedoo.com/youtube/}} downloaded by the \textit{YoutubeStat.py} module \cite{li2018youtubestat}.
\input{table_top20}

\subsection{Text Visualization}\label{sec:visualization}
\input{figure_word_cloud}
Word cloud is a straightforward and appealing visualization method for text. It has served as a starting point for many studies in text mining and opinion mining \cite{heimerl2014word}. Our dataset, which has large amount of text data about people's attitudes toward autonomous driving, is an ideal playground for word cloud generation as well as research in sentimental analysis and opinion mining. Figure \ref{wordcloud} provides word clouds of the dataset generated by using a wordcloud library in Python\footnote{\url{https://github.com/amueller/word_cloud}}. On March 19, 2018, a self-driving Uber vehicle killed a pedestrian in Arizona and this incident is believed to be the first fatality associated with self-driving technology \cite{wakabayashi2018self}. Figure \ref{before} is generated from comments one month before the incident while Figure \ref{after} shows a word cloud of comments one month after. We can see a clear divergence of topics in these two figures where the happening of the incident was the watershed.

\subsection{Sentimental Analysis}\label{sec:sentimental}
Finding out what other people think has always been an important part of information collection and decision making. Gathering opinions of hot, and probably controversial topics in the social media has aroused tremendous attention in the research community. In recent years, autonomous driving technologies have achieved great progresses and gradually become such a topic - popular and controversial. Self-driving cars have been put through millions of miles of road tests, and some believe that the technology has the potential to be safer than human drivers \cite{koopman2017autonomous}. However, with the recent fatal incident, public concerns of safety are clearly increasing according to the time series boxplot in Figure \ref{fig:sentiment_boxplot}, which also shows an interesting process of destroying and rebuilding public trusts over time, illustrated by the boxes that went up to the peak in February, then dropped dramatically in March because of the incident, and finally went up again recently.

Sentiment scores used in the boxplot are floating-point numbers representing sentiment polarities ranging from -1 to 1, where -1 is extreme negative, 0 is neutral, and $1$ is extreme positive. The scores are already given in the dataset and were generated by the sentiment analysis feature of the Google Cloud Natural Language API\footnote{\url{https://cloud.google.com/natural-language/}}. It is natural that some might argue the validity of the API. We admit that the API is not perfect according to our experiments, and will provide a detailed comparison of results from this API and other state-of-the-art models in our upcoming paper. For now, we just introduce a model and leave other concerns to our next paper. The model is
\begin{equation}
    \Phi(\Theta) = \phi(\Theta) + \epsilon_{\phi}
\end{equation}
where $\Theta$ is an embedding of text, $\phi(\cdot)$ is a prediction model, $\Phi(\cdot)$ is the real model to be found, and $\epsilon_{\phi}$ is independent to $\Theta$ but dependent to $\phi$. The $\epsilon_{\phi}$ of the model given by the API is considered to be white noise.

\input{figure_sentiment_boxplot}

\subsection{Recommendation Systems}\label{sec:recommendation}
With the expectation of immense commercial values \cite{graham2007social}, recommendation systems for advertisement have been studied extensively, and two main paradigms have emerged: content-based recommendation and collaborative recommendation \cite{balabanovic1997fab}. Traditional content-based recommendation systems are Information Filtering (IF) systems that need proper techniques for representing the items and producing the user profile, and some strategies for comparing the user profile with the item representation, including content analyzer, profile learner, and filtering component \cite{lops2011content}. The YouTube AV 50K dataset can be used either to generate sentimental scores mentioned in section \ref{sec:sentimental} for user profile enrichments or to directly encode comment texts and metadata as inputs for deep neural networks to finally output ratings in recommendation systems.

\subsection{Text Regression}\label{sec:regression}
Text regression is a method of predicting a real-world continuous quantity associated with the text's meaning based on a piece of text and it was reported to significantly outperform past volatility in predicting financial indices \cite{kogan2009predicting}. Empirically, sales of products could be greatly influenced by online reviews. The case would be especially true for vehicle industry, because vehicles are generally more valuable than normal goods, and intuitively, people would be more careful in the selection processes and tend to look up more reviews. Considering these features, the YouTube AV 50K dataset is appealing for vehicle market as a source of sales prediction and risk measurement research. One of our upcoming papers would combine Bayesian deep learning with text regression to address this issue \cite{li2018modeling, lin2018deep}.

\section{Related Works}\label{sec:related}
The YouTube AV 50K is the first dataset dedicated for social media analysis in autonomous vehicle-related fields. Other very large datasets for social media analysis includes:
\begin{itemize}
    \item Amazon Fine Food Reviews\footnote{\url{https://www.kaggle.com/snap/amazon-fine-food-reviews}}: 568,454 food reviews Amazon users left up to October 2012.
    \item Amazon Reviews\footnote{\url{https://snap.stanford.edu/data/web-Amazon.html}}: Stanford collection of 35 million amazon reviews.
    \item Disasters on social media\footnote{\url{https://www.figure-eight.com/data-for-everyone/}}: 10,000 tweets with annotations whether the tweet referred to a disaster event.
    \item Reddit Comments\footnote{\url{https://www.reddit.com/r/datasets/comments/3bxlg7/i_have_every_publicly_available_reddit_comment/}}: every publicly available reddit comment as of july 2015. 1.7 billion comments.
    \item Twitter Cheng-Caverlee-Lee Scrape\footnote{\url{https://archive.org/details/twitter_cikm_2010}}: Tweets from September 2009 - January 2010, geolocated.
    \item Classification of political social media\footnote{\url{https://www.figure-eight.com/data-for-everyone/}}: Social media messages from politicians classified by content.
    \item Corporate messaging\footnote{\url{https://registry.opendata.aws/commoncrawl/}}: A data categorization job concerning what corporations actually talk about on social media.
\end{itemize}
Above is a selected list and not intended to be comprehensive. More related open datasets can be found in following Github repositories:
\begin{itemize}
\item \textit{niderhoff/nlp-datasets}\footnote{\url{https://github.com/niderhoff/nlp-datasets}}
\item \textit{awesomedata/awesome-public-datasets}\footnote{\url{https://github.com/awesomedata/awesome-public-datasets}}.
\end{itemize}

\section{Future Works}\label{sec:future}
We want to expand the impacts of our previous research in autonomous vehicles \cite{gong2018cooperative, lin2015modeling, li2018modeling, lin2018deep} and have better understanding of what people think of self-driving cars. We are especially interested in people's opinions toward the transitions from level 0: no automation to level 5: full automation \cite{blumberg1995multi, yan2007autonomous}. Although YouTube AV 50K already provides rich sources of data annotated by state-of-the-art NLP techniques, we aware of the shortages of the techiques; for example, doing aspect-level sentimental analysis and detecting sarcasm are still challenging. We would keep an eye on latest progesses in the NLP community and be open and ready to apply such techniques to our dateset.

Another improvement we would like to perform is to keep expanding the dataset as the autonomous driving field is growing rapidly and many exciting achievements are expected, which might greatly change the self-driving landscape as well as public opinions. Hopefully, such changes would mitigate people's security concerns and bring out safer, smoother, and smarter transportation.

\section*{Acknowledgments}
We thank authors of \cite{liu2018mining,li2018semi,wang2018cooperative,zhou2016ultrasonically} for helpful discussions, and the anonymous reviewers for useful feedbacks.

%% file: table_stat.tex
\begin{table}[t]
\begin{center}
\begin{tabular}{|c|c|}
\hline
Item & \# \\
\hline
Channels & 50 \\
Data size (MB) & 119 \\
Videos & 632 \\
Annotated comments & 19,126 \\
Unique authors & 21,005 \\
First level comments & 21,498 \\
Total comments & 30,456 \\
Word count & 796,371 \\
\hline
\end{tabular}
\end{center}
\caption{Basic statistics of YouTube 50K.}
\label{tab:stat}
\end{table}

%% file: figure_comment_json.tex
\begin{figure}[!t]
\centering
\includegraphics[width=2.0in]{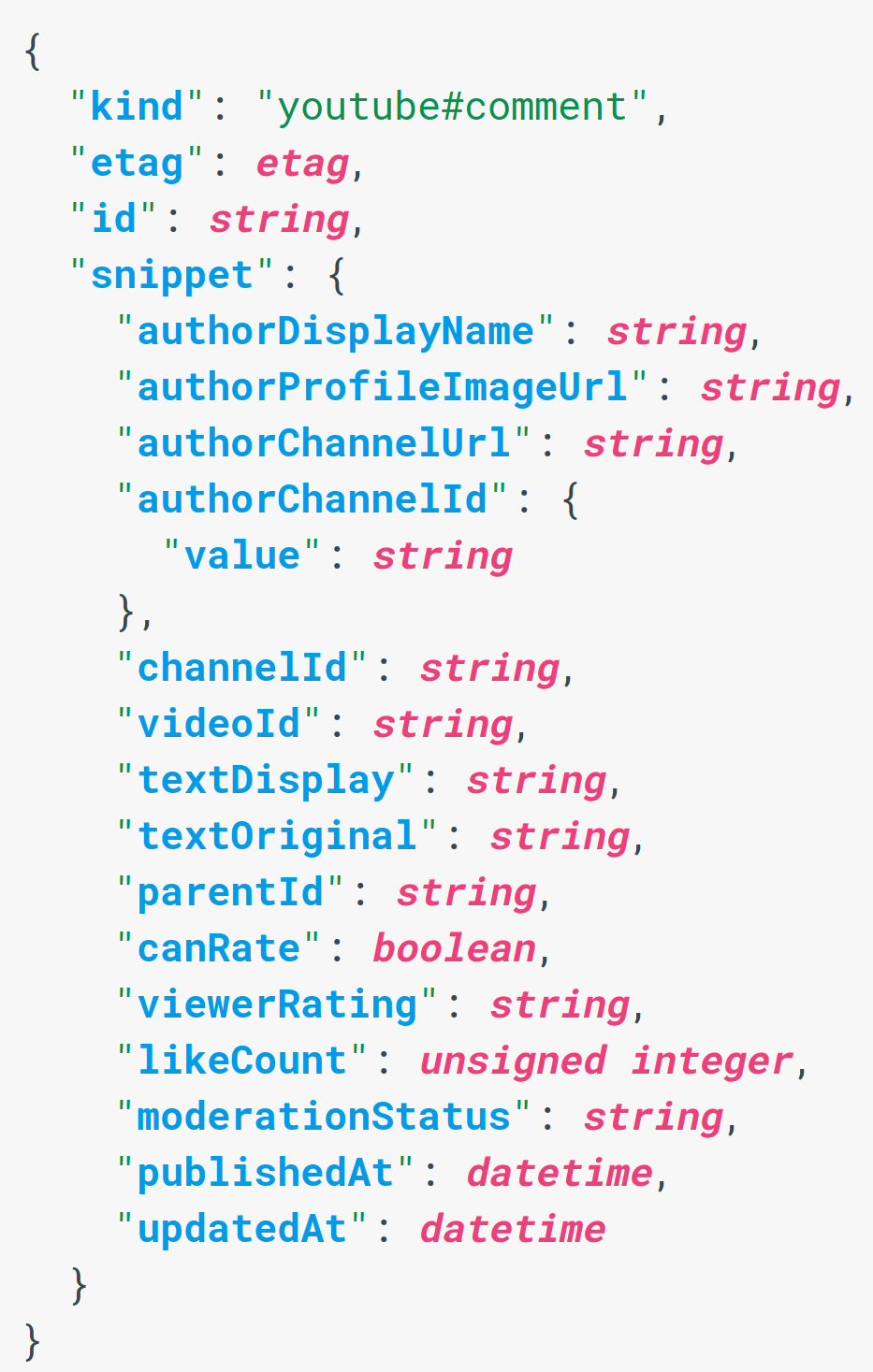}
% where an .eps filename suffix will be assumed under latex,
% and a .pdf suffix will be assumed for pdflatex; or what has been declared
% via \DeclareGraphicsExtensions.
\caption{JSON structure shows the format of a comments resource.}
\label{fig:comment_json}
\end{figure}

%% file: figure_meta_histgram.tex
\begin{figure}[!t]
\centering
\subfloat[number of comments by hour]{\includegraphics[width=3in, height=2.3in]{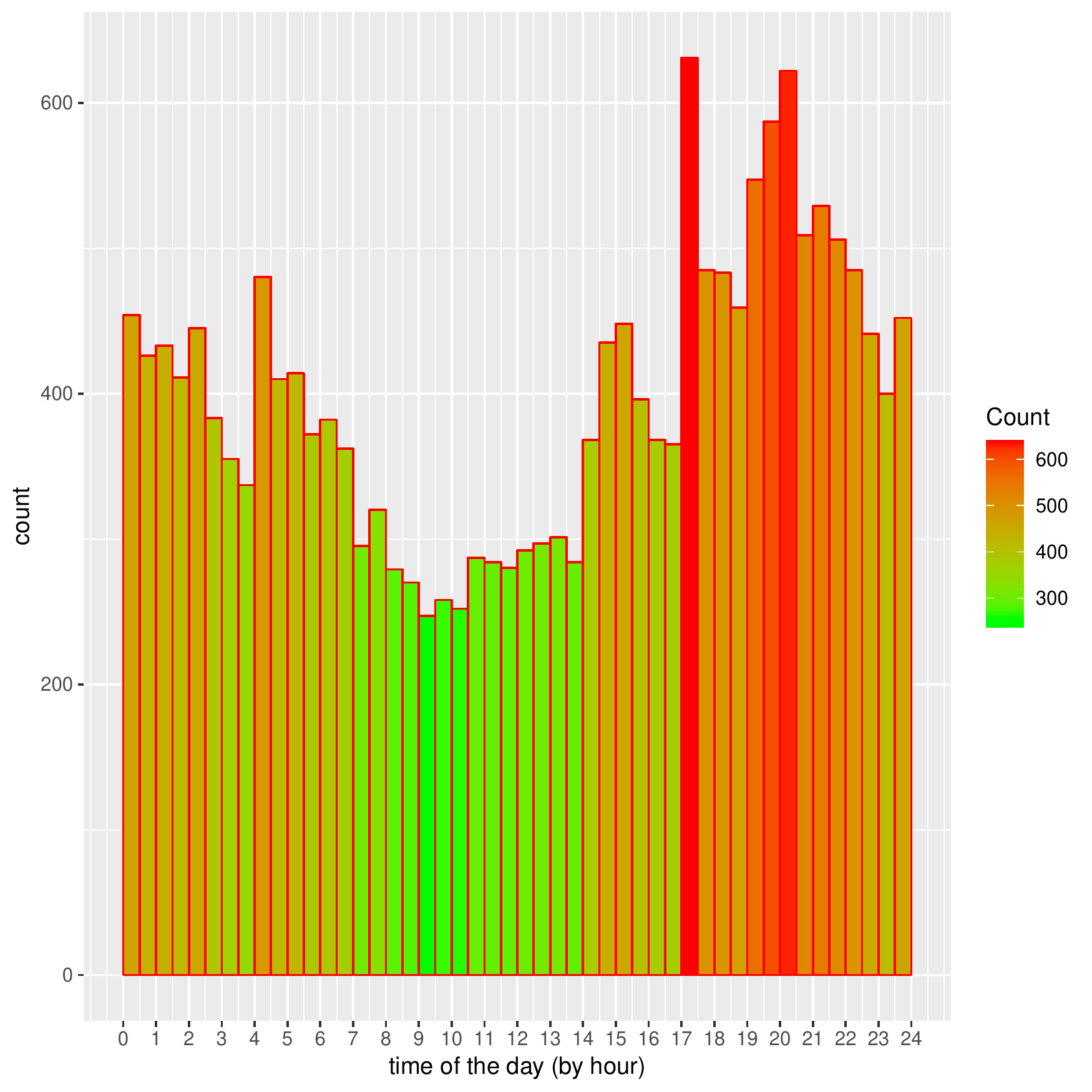}
\label{aasdf1}}
\hfil
\subfloat[number of comments by weekday]{\includegraphics[width=3in, height=2.3in]{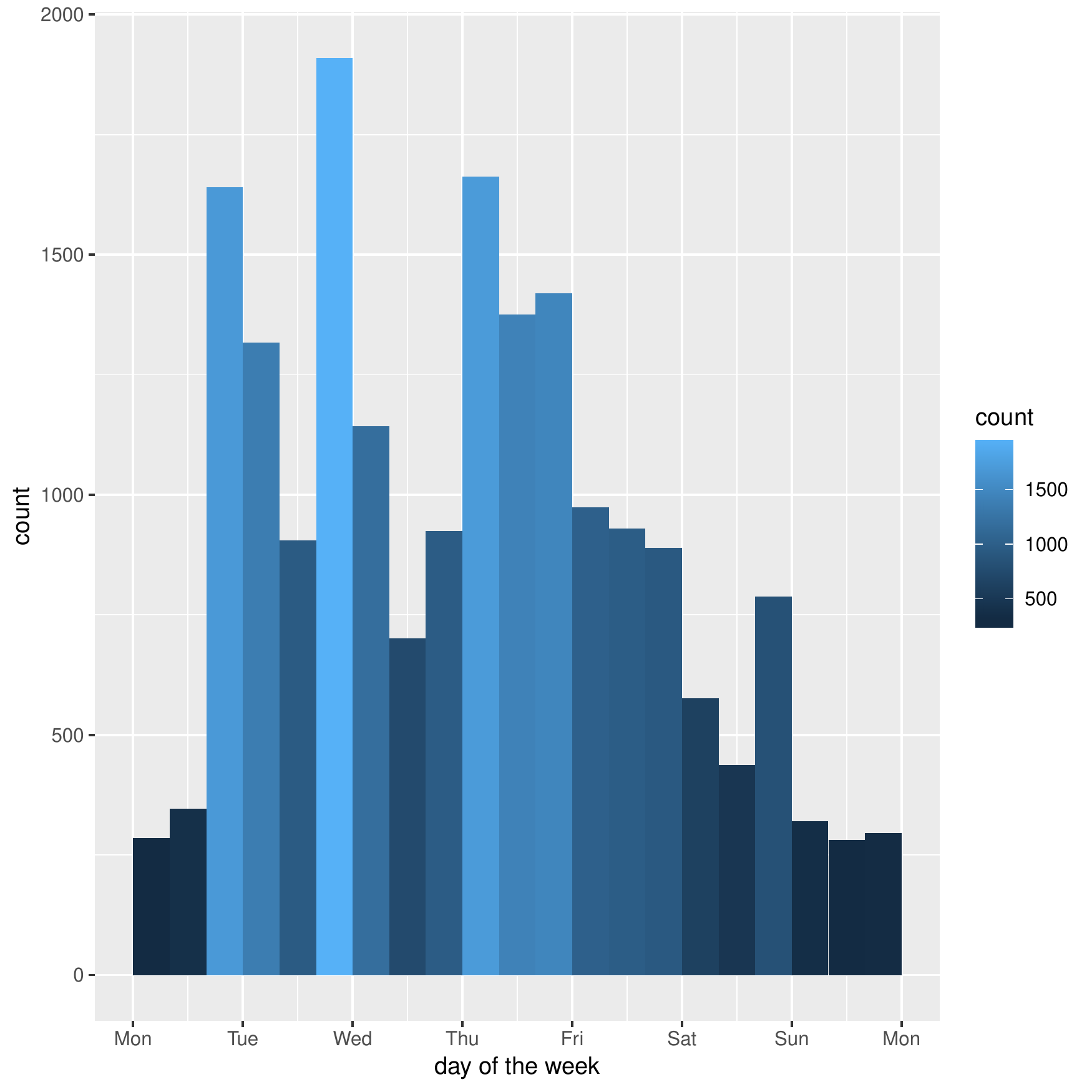}
\label{2adsf}}
\caption{Distribution of comment publish time.}
\label{fig:meta_histgram}
\end{figure}

%% file: table_top20.tex
\begin{table}
\begin{center}
\begin{tabular}{|c|c|c|c|}
\hline
\# & Channel & Monthly Views & Subscribers \\
\hline
1 & Inside Edition & 395025524 & 3129845 \\
2 & CNN & 155319328 & 3794747 \\
3 & ABC News & 144872445 & 3850158 \\
4 & Barcroft TV & 129030102 & 4627820 \\
5 & MSNBC & 74032667 & 963021 \\
6 & Fox News & 60536948 & 1242054 \\
7 & Crime Watch Daily & 53485050 & 877800 \\
8 & BBC News & 45344148 & 2619048 \\
9 & The Young Turks & 45159957 & 3899180 \\
10 & Guardian News & 42598679 & 136110 \\
11 & RT & 40166881 & 2632872 \\
12 & TODAY & 40073086 & 758845 \\
13 & The Royal Family Channel & 38726120 & 314798 \\
14 & USA TODAY & 38444609 & 634366 \\
15 & VICE News & 36201734 & 3048964 \\
16 & TomoNews US & 34828513 & 2230927 \\
17 & CBS News & 34513896 & 752014 \\
18 & DramaAlert & 34501164 & 3888026 \\
19 & CBS This Morning & 34236213 & 390538 \\
20 & Vox & 32228568 & 4166051 \\
\hline
\end{tabular}
\end{center}
\caption{Top 20 YouTube English news channels ranked by monthly reviews (from May 9th, 2018 to June 9th, 2018)}
\label{table_top20}
\end{table}

%% file: figure_word_cloud.tex
\begin{figure*}[t]
\centering
\subfloat[one month before the crash]{\includegraphics[width=3.1in,frame]{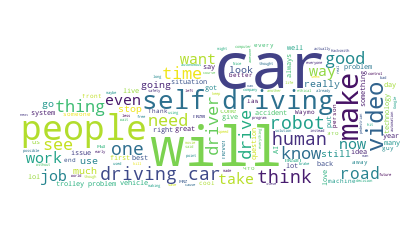}
\label{before}}
% \hfil
% \subfloat[the fatal crash scene]{\includegraphics[width=2.8in]{uber_arizona_crash.jpg}
% \label{crash}}
% \hfil
\subfloat[one month after the crash]{\includegraphics[width=3.1in]{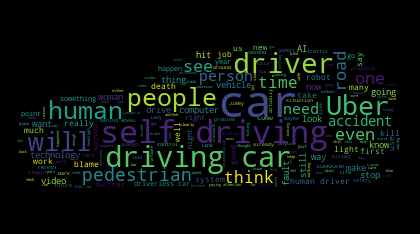}
\label{after}}
\caption{Word cloud generated from YouTube comments below self-driving related videos before and after the first self-driving car fatality on March 19, 2018 in Tempe, Arizona.}
\label{wordcloud}
\end{figure*}

%% file: figure_sentiment_boxplot.tex
\begin{figure}[!t]
\centering
\includegraphics[width=2.5in]{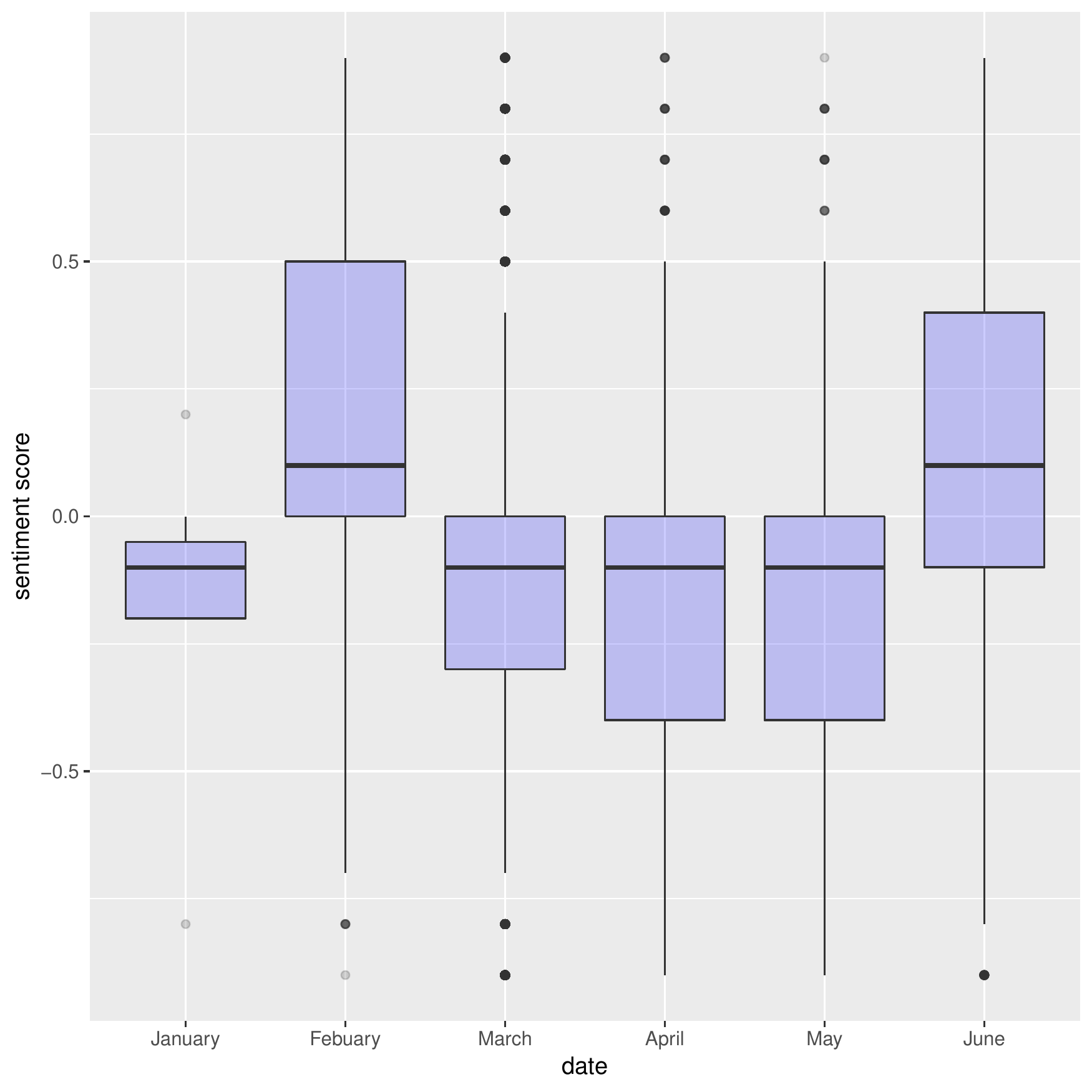}
\caption{Boxplot of sentiment scores changes over time.}
\label{fig:sentiment_boxplot}
\end{figure}